\documentclass[letterpaper, 10 pt, conference]{ieeeconf}  

\IEEEoverridecommandlockouts                              

\usepackage{graphics} 
\usepackage{epsfig} 
\usepackage{mathptmx} 
\usepackage{times} 
\usepackage{amsmath} 
\usepackage{amssymb}  
\usepackage{multirow}
\usepackage{subfigure}

\title{\LARGE \bf
A Low-Cost Tele-Presence Wheelchair System*
}

\author{Jiajun Shen$^{\dag}$, Bin Xu$^{\dag}$, Mingtao Pei, Yunde Jia
\thanks{*Resrach supported in part by the Natural Science Foundation of China (NSFC) under Grant No.61375044.}
\thanks{$^{\dag}$The two authors contributed equally to this work.}%
\thanks{All authors are with the School of Computer Science, Beijing Institute of Technology, and Beijing Lab of Intelligent Information Technology, Beijing 100081, CHINA. {\tt\small E-mail: \{shenjiajun, xubinak47, peimt, jiayunde\}@bit.edu.cn}}%
}

\begin{document}

\maketitle
\thispagestyle{empty}
\pagestyle{empty}

\begin{abstract}

This paper presents the architecture and implementation of a tele-presence wheelchair system based on tele-presence robot, intelligent wheelchair, and touch screen technologies. The tele-presence wheelchair system consists of a commercial electric wheelchair, an add-on tele-presence interaction module, and a touchable live video image based user interface (called TIUI). The tele-presence interaction module is used to provide video-chatting for an elderly or disabled person with the family members or caregivers, and also captures the live video of an environment for tele-operation and semi-autonomous navigation. The user interface developed in our lab allows an operator to access the system anywhere and directly touch the live video image of the wheelchair to push it as if he/she did it in the presence. This paper also discusses the evaluation of the user experience.

\end{abstract}

\section{INTRODUCTION}

\begin{figure}[t]
\setlength{\abovecaptionskip}{-8.px}
\begin{center}
    \includegraphics[width=0.45\textwidth]{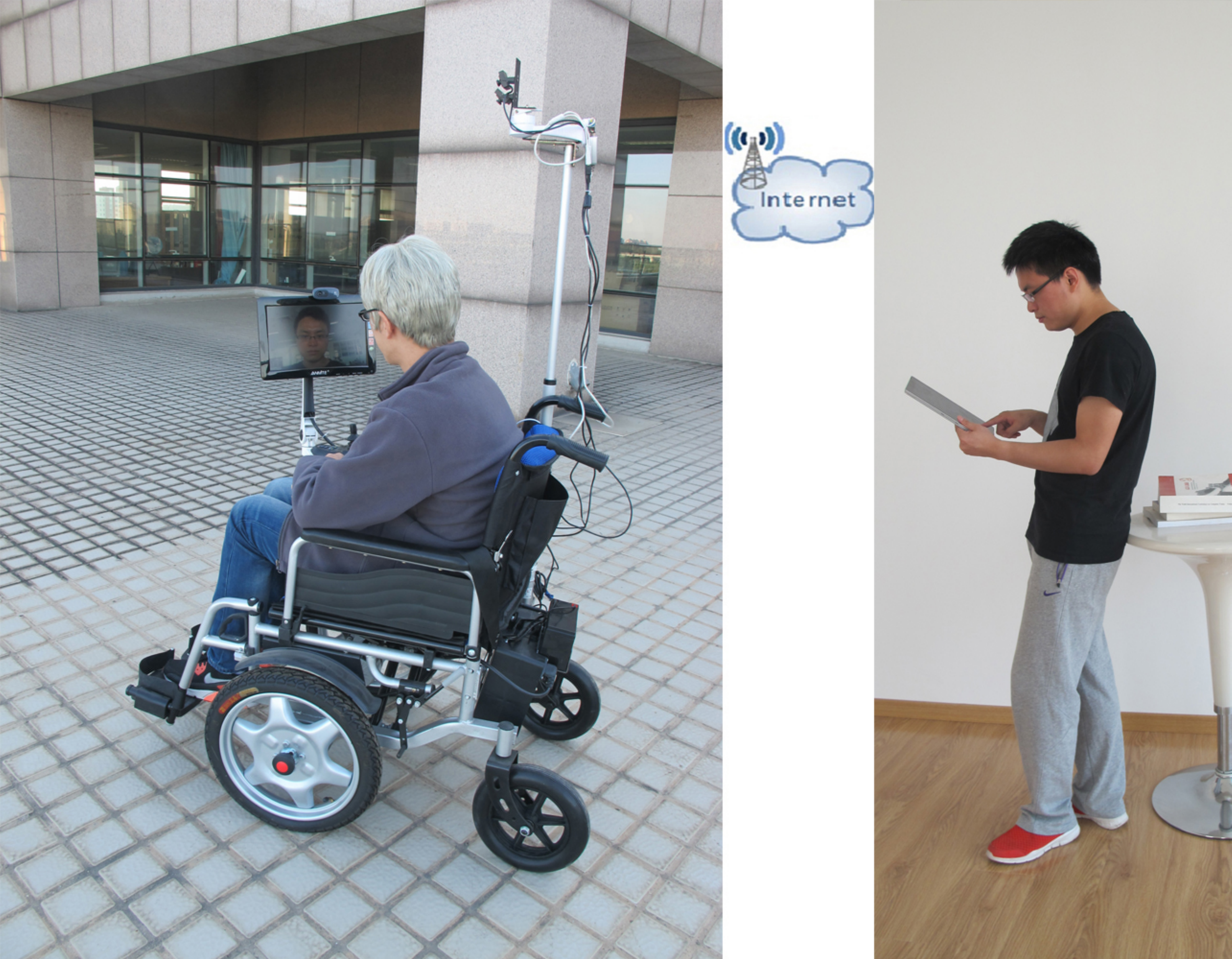}
\end{center}
\caption{Our tele-presence wheelchair system. The operator in a remote location uses a pad to tele-operate (push) the tele-presence wheelchair in which an elderly person is sitting, while the elderly is video-chatting with the operator (family member or caregiver) on the screen in front of the chair.}
\label{fig:fig1}

\end{figure}

Wheelchairs are powerful assistant devices for disabled and elderly people to have mobility. Wheelchairs have evolved from manual wheelchairs (MW) to electric wheelchairs (EW) and intelligent wheelchairs (IW) (or robotic wheelchair, smart wheelchair) \cite{bedaf2015overview}\cite{cheng2011development}. Intelligent wheelchairs improve the traditional wheelchair features with the navigation capability and automatic adaptation of interfaces to operators or users \cite{bedaf2015overview}\cite{cheng2011development}\cite{simpson2005smart}. Intelligent wheelchairs commonly have three operation modes: manual mode, electric mode, and intelligent mode. It is easy to switch among the three modes for different users and different cases. In this paper, we present a tele-presence wheelchair (TW) which is working in the fourth operation mode, called tele-presence mode. This mode allows an operator (family member or caregiver) to use a pad to operate the wheelchair in a remote location as if he/she did it in the presence, Fig.~\ref{fig:fig1} shows our tele-presence wheelchair(TW). An operator uses a pad in a remote location to operate the tele-presence wheelchair in which an elderly is sitting, while the elderly is video-chatting with the operator (family member or caregiver) on the screen in front of the wheelchair.

Our work is motivated by two requirements that are crucial for elderly and disabled people. Firstly, elderly and disabled people have a limited capability of controlling the wheelchair, so extra assistance is needed. To reduce the difficulty of operation, intelligent wheelchairs have received considerable attention in robotics and artificial intelligent communities. But most projects used expensive devices and sophisticated technologies to perform autonomous navigation and adaptable interface with high cost and heavy training, which blocks the intelligent wheelchair to move into commercialization phrase. Secondly, geriatric depression is widely spread in the group of elderly people, so the company of family members is extremely crucial to them. Tele-presence robots have been successfully used in accompanying elderly and disabled people to alleviate loneliness problem and play etiological role in physical and mental health problems \cite{hicks2000your}. But existing tele-presence robots do not offer the assistance of mobility for elderly and disabled people. We add a tele-presence interaction hardware module on an electric wheelchair to combine the advantages of intelligent wheelchairs and tele-presence robots with very low cost and high safety. Our tele-presence wheelchair can be not only the embodiment of family members to accompany elderly and disabled people, but also assist the mobility of them.

This paper presents the architecture and implementation of the tele-presence wheelchair system based on the technologies supporting to tele-presence robots, intelligent wheelchairs, and touch screens. The tele-presence wheelchair is a commercial electric wheelchair equipped with a tele-presence interaction module. The tele-presence interaction module is used to provide video-chatting for an elderly or disabled person with the family members or caregivers, and also capture the live video of an environment for tele-operation and semi-autonomous navigation. The user interface developed in our lab is the TIUI \cite{jia2015telepresence}, which is a touchable live video image based user interface for a smart pad. The TIUI allows an operator to access the system anywhere to directly touch and push the live video images of a tele-presence wheelchair in a remote location as if the operator did it in the presence..

The tele-presence wheelchair can be easily tele-operated by any user, especially novice users. It is a good solution of the physical limitation of presence i.e. one person cannot be present in two places at the same time \cite{kaufmann2014toward}.

\section{RELATED WORK}

Our world is facing problems associated with an increasing elderly population. It was found that activities concerning mobility, self-care, interpersonal interaction and relationships are most threatening with regard to the independent living of elderly people \cite{lehmann2013should}. In order to maintain the quality of home care for the elderly, assistive robots and other technologies have been given increased attention to support the care and independence of elderly people \cite{bedaf2015overview}. Our wheelchair is one of such robots. It is related to tele-presence robot and intelligent wheelchair technologies.

\subsection{Tele-presence robots}

A tele-presence robot is a mobile robot incorporating a video conferencing device installed on it and provides a more flexible tele-presence experience by allowing participants to have some degree of mobility in a remote environment \cite{kristoffersson2013review}.

One of the important application domains for tele-presence robots is elderly care and health care, where tele-presence robots can be profitable and contribute to the prevention of problems related to loneliness and social isolation \cite{coradeschi2011towards}.

Boissy et al. \cite{boissy2007qualitative} presented the concept of a tele-presence robot in home care for elderly people in 2007, and their qualitative research identified potential applications where elderly people might use such a robot, such as to connect with family members. Tsai et al. \cite{tsai2007developing} also developed a tele-presence robot to allow elderly people to communicate with the family members or caregivers. They found that the tele-presence robot enabled elderly people to regard the tele-presence robot as a representation of the robot operator.

Nowadays, commercial available tele-presence robots include the Giraff \cite{Giraff} and the VGo \cite{VGo}, designed specifically for elderly people. There are many other general tele-presence robots available to provide facilities in nursing homes and health care centers \cite{kristoffersson2013review}. The mobility is one of the most important daily life activities of elderly or disabled people, but so far, none of tele-presence robots take into account the mobility, and offer the assistance of mobility for elderly or disabled people. Another critical issue is the interface which is typically designed for tele-operation on desktop computer with keyboard or mouse and requires a highly trained operator. Such interfaces can not allow a user to have convenient access to the system to tele-operate the robot via smart mobile devices, such as pads and smart phones popularly used today.

\subsection{Intelligent wheelchairs}

Intelligent wheelchairs (IW) or smart wheelchairs to help the mobility of elderly or disabled people were introduced in the 1980s \cite{cheng2011development}\cite{madarasz1986design}. Simpson et al. \cite{simpson2004smart} and Faria \cite{faria2014survey} provide two comprehensive reviews of intelligent wheelchairs. Typically, an IW is controlled by a computer which can perform the perception of the environments around the wheelchair through many sensors by using intelligent algorithms \cite{simpson2004smart}.

User-machine interface and autonomous navigation are two of the most important techniques in developing intelligent wheelchairs. The interface consists of not only a conventional wheelchair joystick, but also controls based on voice, facial expressions, gaze, body action, and multimodal perception \cite{Giraff}. Another emerging interface is brain-based controls and has received a significant attention \cite{simpson2004smart}. Autonomous navigation mainly ensures the wheelchair's safety, flexibility, and obstacle avoidance capabilities based on many sensors. Most autonomous navigation techniques of intelligent wheelchairs have been derived from autonomous robot technologies. Recent work includes wheelchair navigation based on artificial intelligence (AI) and advanced computing technologies \cite{henry2010learning}\cite{morales2013human}, obstacle-avoidance \cite{cheng2011development}, and automatic target tracking \cite{carlson2010increasing}. But so far there are few intelligent wheelchairs commercial available and still need long time to resolve the limitations and challenges such as the adaption, safety and cost, especially expensive sensors and complicated environments \cite{henry2010learning}\cite{kim2015robotic}\cite{tsui2015accessible}.

Our tele-presence wheelchair is an affordable commercial electric wheelchair equipped with a low-cost tele-presence interaction module. It can offer the assistance of mobility for elderly or disabled people, and can be an embodiment of family members or caregivers to accompany them.

\section{SYSTEM OVERVIEW}
\begin{figure}[t]
\setlength{\abovecaptionskip}{-8.px}
\begin{center}
    \includegraphics[width=0.45\textwidth]{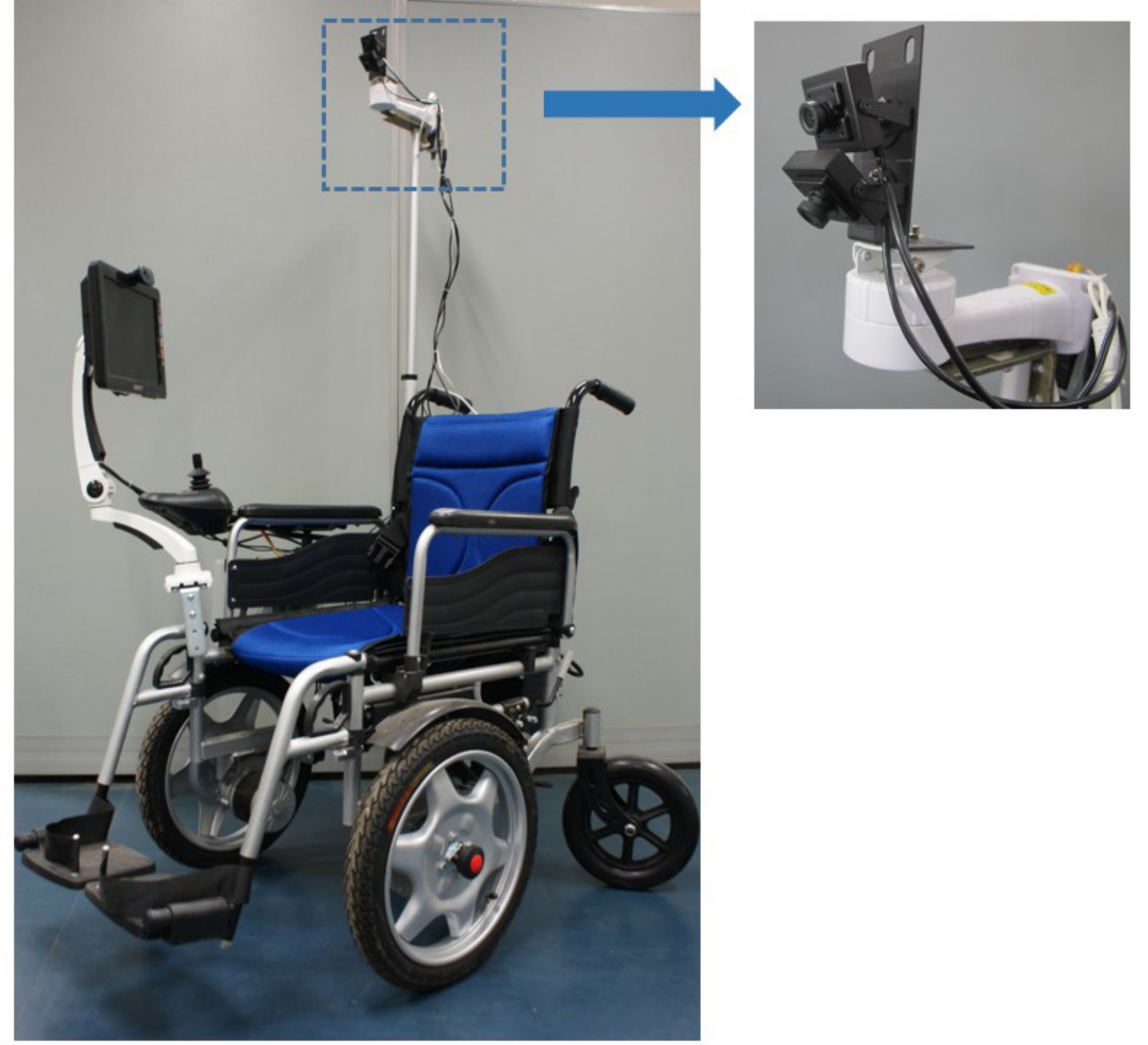}
\end{center}
\caption{The prototype of our tele-presence wheelchair. On the top right is the close-up image of the tele-operation imaging part.}
\label{fig:fig2}
\end{figure}

\begin{figure}[t]
\setlength{\abovecaptionskip}{-8.px}
\begin{center}
    \includegraphics[width=0.45\textwidth]{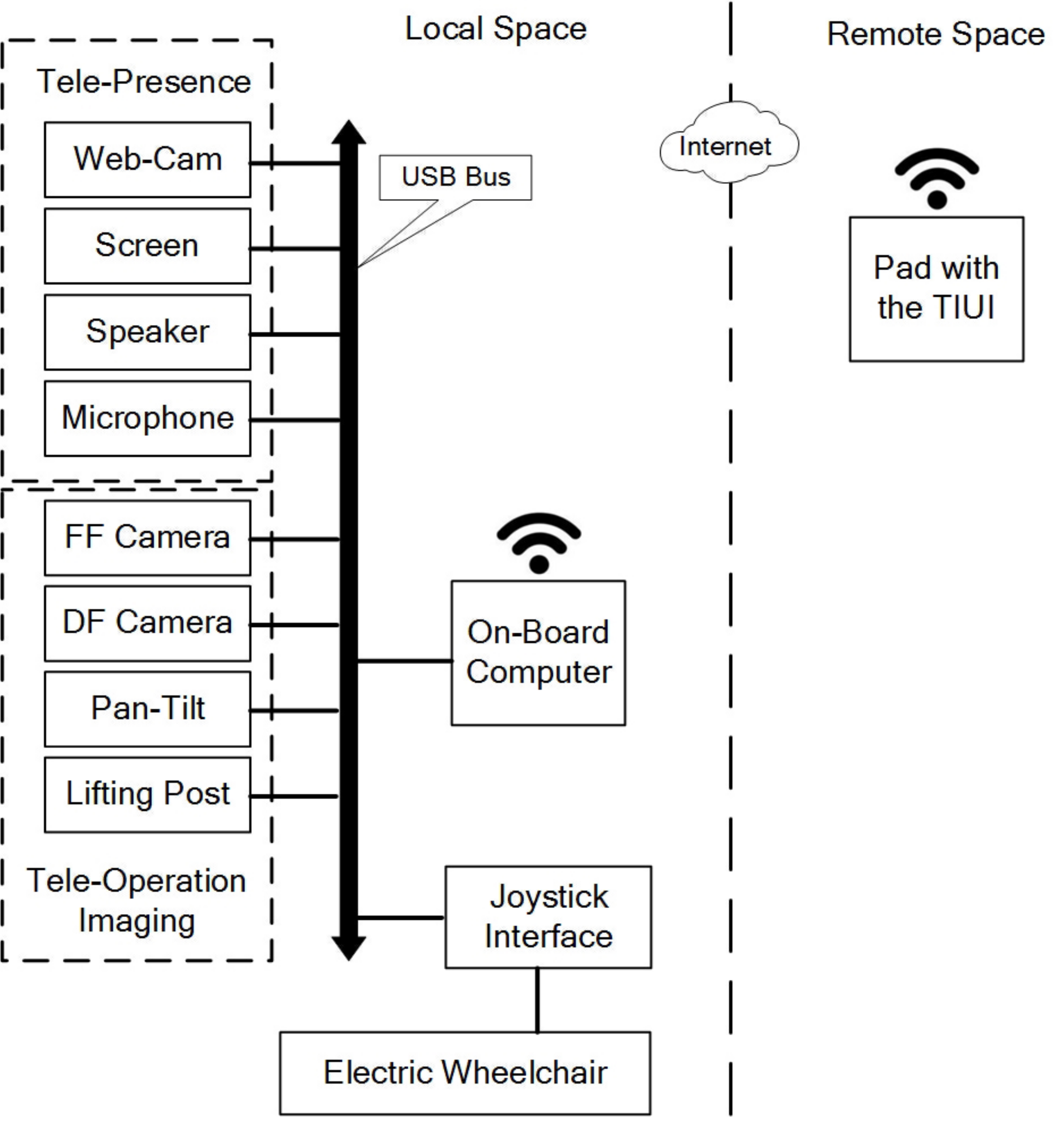}
\end{center}
\caption{The architecture of the tele-presence wheelchair system}
\label{fig:fig3}
\end{figure}

Fig.~\ref{fig:fig2} shows the prototype of the tele-presence wheelchair developed in our lab. Fig.~\ref{fig:fig3} shows the architecture of the tele-presence wheelchair system. Following our previous work \cite{jia2015telepresence}, we define a space as the local space in which the wheelchair moves, a space as the remote space in which an operator uses the TIUI to operate the wheelchair, and connect the two spaces by wireless internet communication.

The tele-presence interaction module contains three parts: tele-presence, tele-operation imaging, and user interface. The tele-presence part is mounted on the front of the wheelchair for an elderly or disabled person to video-chat with family members. This part is composed of a Web-Cam, a microphone, a speaker, and a light LCD screen. It enables a disabled or elderly person to video-chat with his/her family members. The tele-operation imaging part uses a forward-facing camera (FF camera) to capture the live video of a local space for the operator to recognize objects ahead for tele-interaction, and a down-facing camera (DF camera) to capture the live video of the ground around the wheelchair for the operator to tele-operate the wheelchair. Both the FF camera and DF camera are mounted on the Pan-Tilt platform holding on to the vertical lifting post.

Existing systems often use two live video windows from the FF camera and DF camera, respectively, as visual feedback \cite{lee2011now}\cite{neustaedter2016beam}. We found in our testing system that two or more windows would introduce some confusion over the local space, which makes a remote operator to feel missing some views and the context of the local space, and distracting to switch attention between the two windows and adapt the different windows. Fortunately, the two images of the local space from the two cameras are overlapping and can be easily stitched as one image, for producing one live video streaming and displaying in one window. We call such a stitched image the FDF (Forward-Down-Facing) live video image. The on-board computer performs live video acquisition through the FF camera and DF camera, and stitches them for the TIUI.

\section{USER INTERFACE}

The user interface of the tele-presence wheelchair is located in a remote space where an operator uses a pad or tablet to tele-operate the wheelchair. An FDF live video image on the TIUI can be divided into the upper part from the FF camera and the lower part from the DF camera. It implies that in a live video image, the content of the upper part is focusing on objects ahead for tele-interaction, and the lower part is focusing on the ground for navigation.

\begin{figure}[t]
\setlength{\abovecaptionskip}{-8.px}
\begin{center}
    \includegraphics[width=0.45\textwidth]{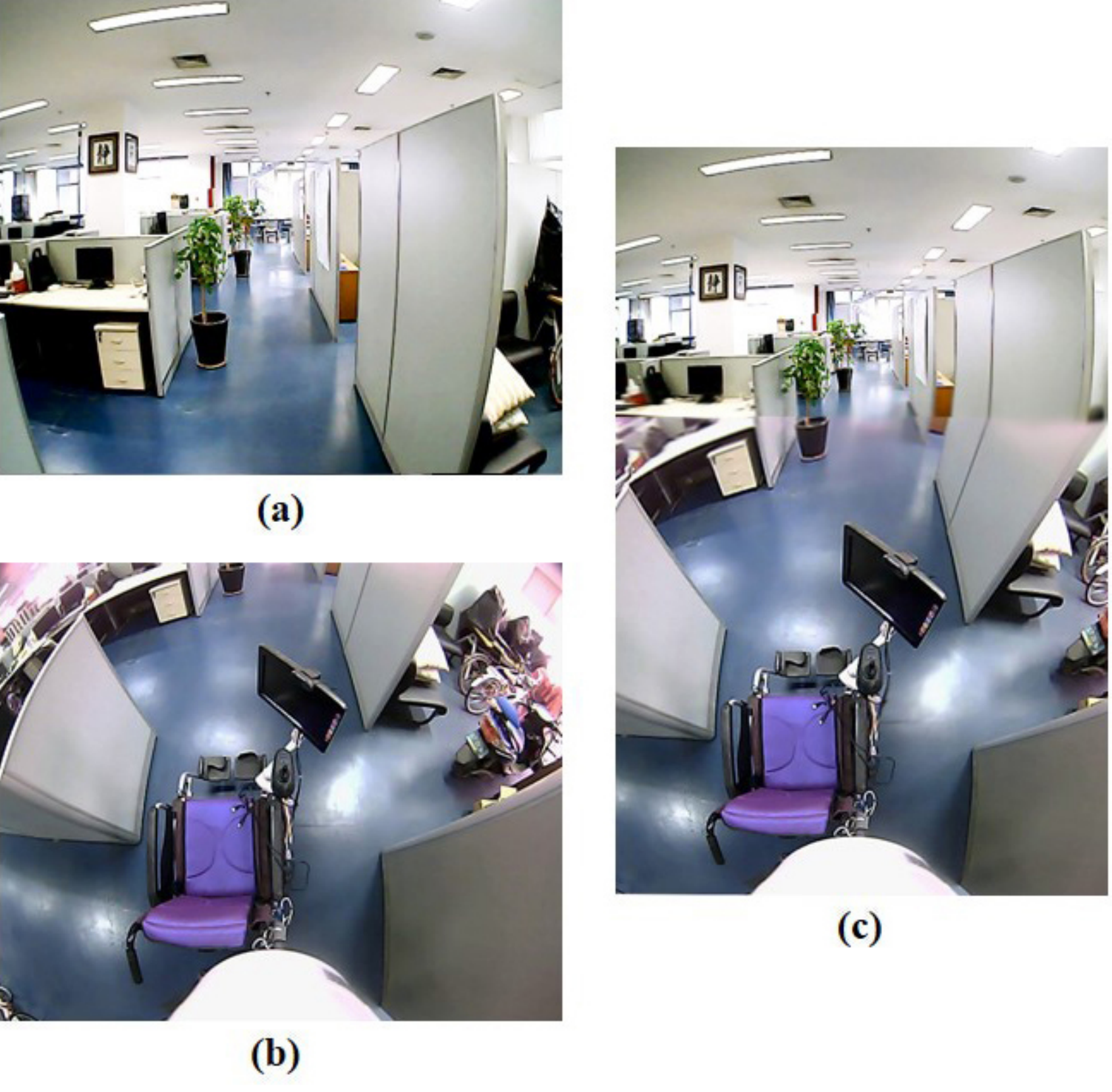}
\end{center}
\caption{Illustration of a live video image on the TIUI (c) by stitching F-F camera image (a) and D-F camera image (b)}
\label{fig:fig4}
\end{figure}

The DF camera uses a very wide angle lens to acquire rich navigation information around the wheelchair. Fig.~\ref{fig:fig4} demonstrates an example of the FDF live video image produced by stitching the two images from the two cameras. Fig.~\ref{fig:fig4} (a)-(b) show the live video images captured by the FF camera and DF camera, respectively. We use a similar method as \cite{brown2007automatic} to stitch the two images, and use SURF instead of SIFT to do the feature matching since SURF is more efficient than SIFT. Fig.~\ref{fig:fig4} (c) shows the stitched image.

\begin{figure}[t]
\setlength{\abovecaptionskip}{-8.px}
\begin{center}
    \includegraphics[width=0.45\textwidth]{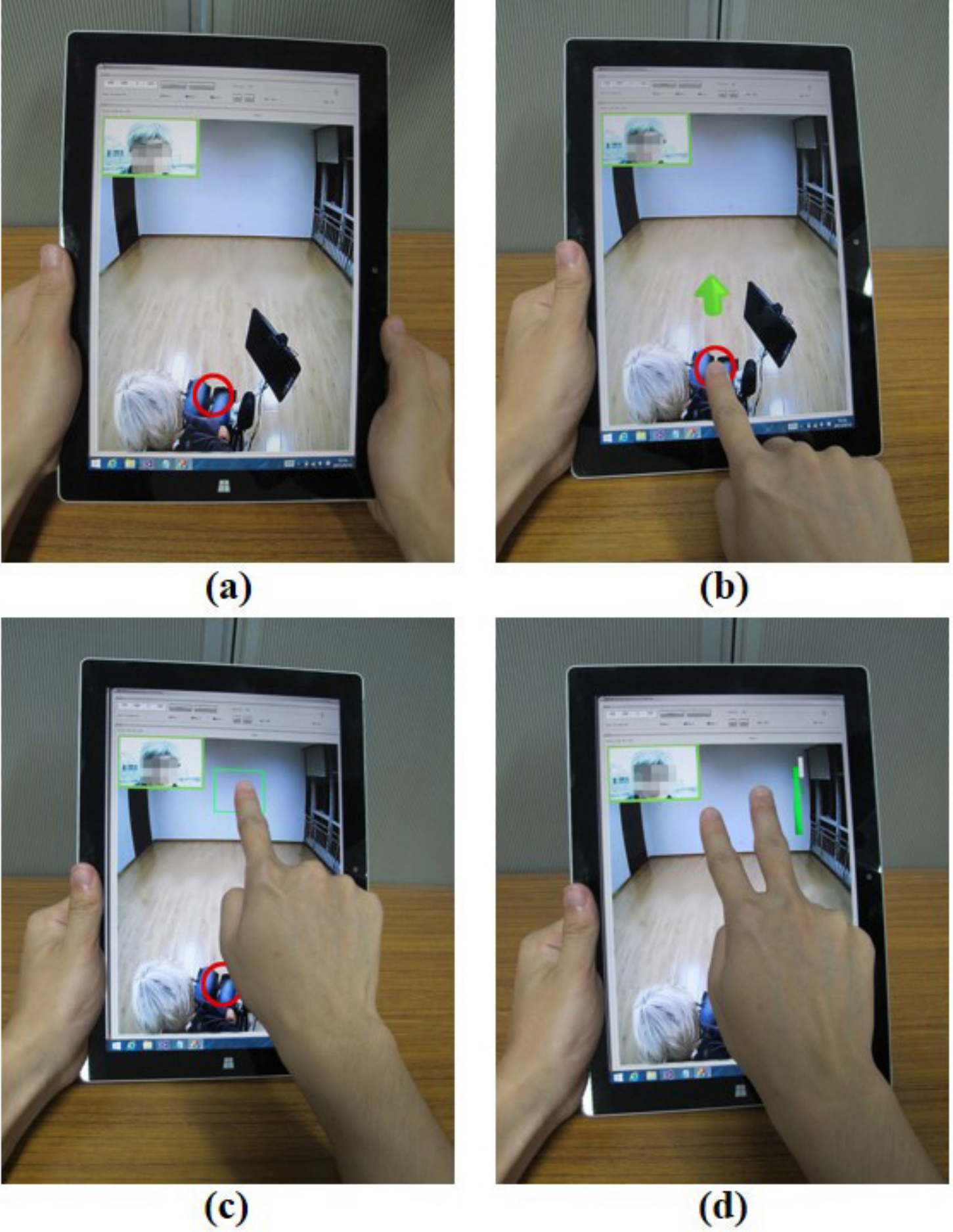}
\end{center}
\caption{Finger touch gestures on the TIUI for tele-operation. (a) The TIUI of our system where the red circle is virtual "track point" of the wheelchair. (b) One-finger touch gesture on the lower part to push the wheelchair to move. (c) One-finger touch gesture on the upper part for tele-interaction with the objects of the local space. (d) Two-finger touch gesture on the upper part to control the motion of the tele-operation imaging device.}
\label{fig:fig5}
\end{figure}

Through the TIUI, we use finger touch gestures to operate the wheelchair and to interact with the environment, as shown in Fig.~\ref{fig:fig5}. In our daily life, we often use our one-finger or two-finger to directly operate almost all the device panels or interfaces as most of them contain switches, buttons, and/or sliders. A joystick can be regarded as a combination of multiple buttons or a track-point of a ThinkPad computer. So, in our system, we also use one-finger and two-finger touch gestures to operate most common devices of daily life in a remote space.

\begin{itemize}

\item We use one-finger touch gestures on the lower part of the live video image (Fig.~\ref{fig:fig5} (b)) to push the wheelchair to move forward/backward, and turn left/right, where the red circle on the low part is a virtual "track point" of the wheelchair, similar to the joystick of a commercial electric wheelchair.
\item We use one-finger touch gestures on the upper part of the live video image (Fig.~\ref{fig:fig5} (c)) to interact with the objects of a local space, such as doors, elevators or vehicles.
\item We use two-finger touch gestures to control the tele-operation imaging device (Pan-Tilt cameras) of the wheelchair to look around or lift up and down to change the height of the camera (Fig.~\ref{fig:fig5} (d)).

\end{itemize}

The gestures are simple and natural. They are easily understood and performed by any users, including novice users.

\section{SEMI-AUTONOMOUS NAVIGATION}

Driving a remotely operated robot is a challenging task and overloading work. To take a specific example of the wheelchair turning left to a doorway. A remote operator can use the one-finger touch gesture to drive the wheelchair turning left, but the wheelchair might turn left too much or too little, then the operator has to use the one-finger touch gesture to adjust the moving direction of the wheelchair. This process may be repeated several times to reach the correct destination, which will make the user experience in the remote space bad and the local person in the wheelchair uncomfortable. The existing smart robot systems are attempting to create a fully autonomous solution with optimal decisions based on position and speed. But their designs are complicated with high cost sensors and computation, and the operator plays an insignificant role in decisions. We develop a human-wheelchair collaborative control algorithm which is user-centered. We call it a semi-autonomous navigation of the wheelchair.

\begin{figure}[t]
\setlength{\abovecaptionskip}{-8.px}
\begin{center}
\subfigure []
{\includegraphics[width=0.45\textwidth]{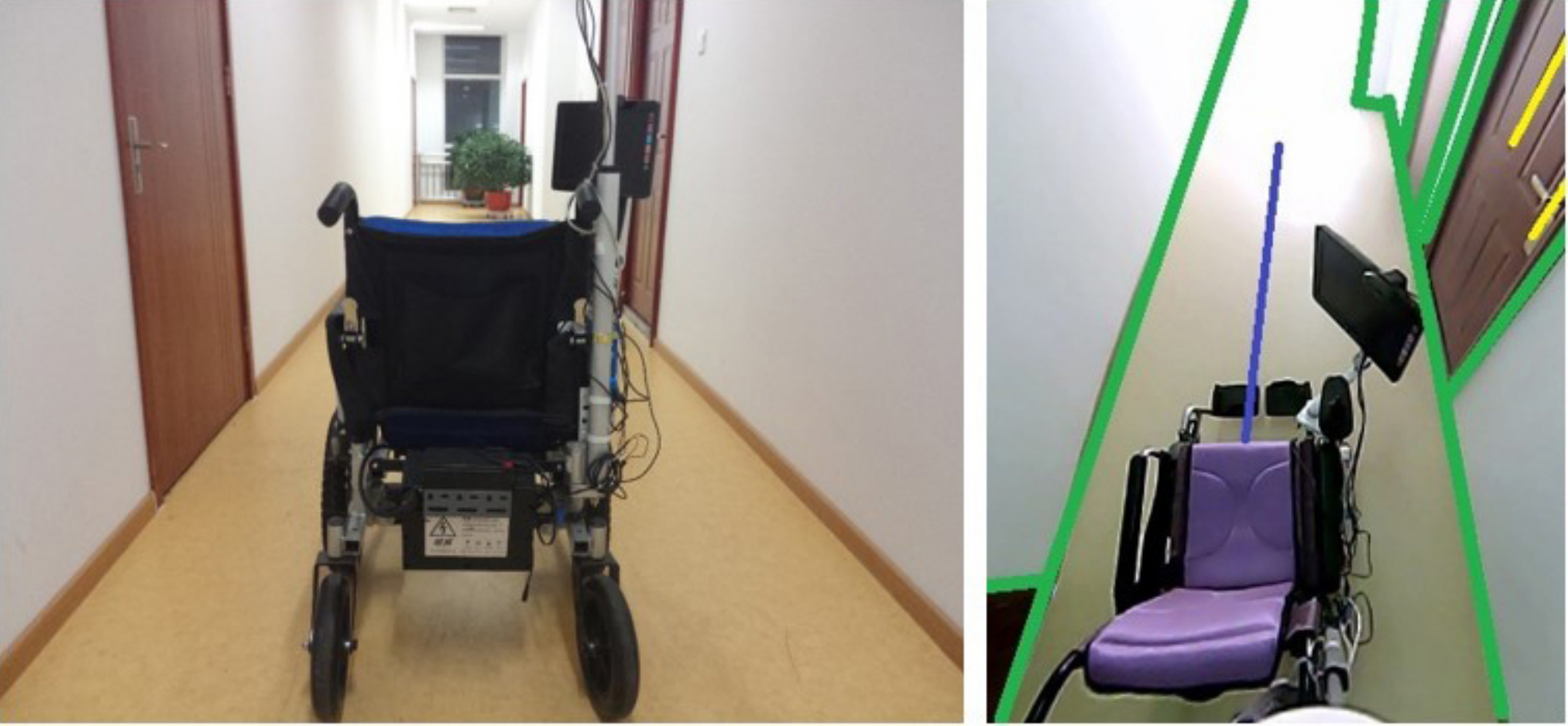}}
\subfigure []
{\includegraphics[width=0.45\textwidth]{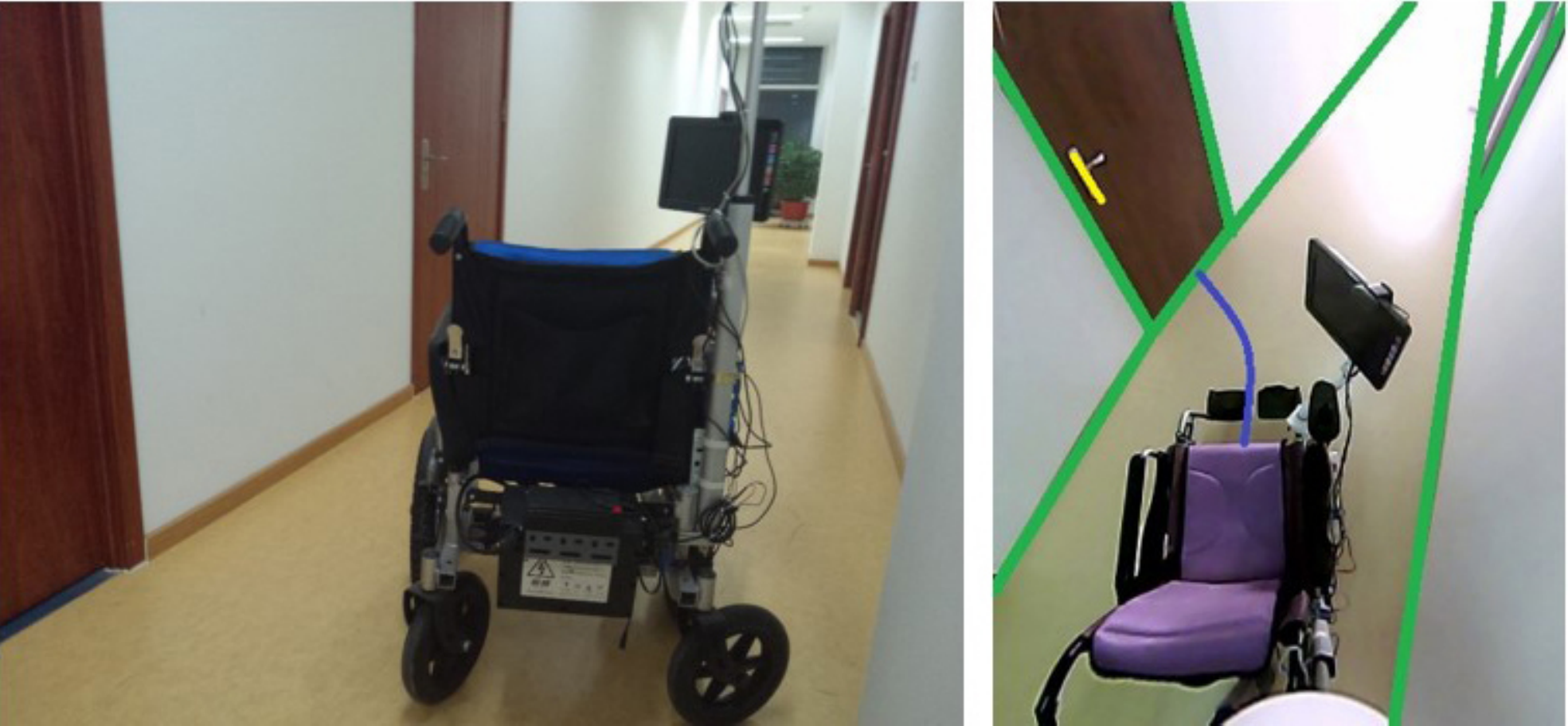}}
\end{center}
\caption{(a) A wheelchair moving forward in the corridor. (b) A wheelchair turning left to enter a door.}
\label{fig:fig6}
\end{figure}

The semi-autonomous navigation assists the operator to push the wheelchair efficiently and gives the person in the wheelchair a good experience. Firstly, the tele-presence wheelchair perceives the environment via live video images without any other high cost sensors. In this part, the corridor corners and doors are detected from live video images by combining edge and corner features \cite{yang2010robust} with some other priori knowledge including vanishing point and the door structure. The door may not be seen completely in the image, so the door-frames and door corners on the ground are the most concerned parts. As shown in Fig.~\ref{fig:fig6} (a), the wheelchair can detect and track the corners and doors on both sides when it is moving along a corridor.

The driving commands contain two basic operations: driving forward and turning left/right. In the driving-forward operation, an operator uses the one-finger touch gesture to push the wheelchair to move forward smoothly as it keeps a safe distance to the corridor corners automatically. In the turning left/right operation, an operator uses the one-finger touch gesture to push the wheelchair to turn left or right. The system recognizes the operator's intention and generates a safe trajectory using the B¨¦zier Curve \cite{bezier1972numerical} according to the user's input and live video images. For example, as shown in Fig.~\ref{fig:fig6} (b), an operator just needs to use the one-finger touch gesture to push the wheelchair to move along the trajectory without caring about the turning angle. If the operator does not want to enter the door, he/she just needs to use another gesture like "turn right" or "move forward", the wheelchair will recognize the operator's new intention and re-generate the trajectory.

\section{EVALUATION}

\subsection{Evaluation of user experience}

We evaluated the user experience of our system by comparing with the situation that operators push the wheelchair in the presence. We recruited 20 volunteers (ages: 18-25 years) to participate in the experiment. They were divided into two equal groups, a remote group to operate/tele-operate the wheelchair and a local group to sit in the wheelchair.

\begin{figure}[t]
\setlength{\abovecaptionskip}{-8.px}
\begin{center}
\subfigure []
{\includegraphics[width=0.45\textwidth]{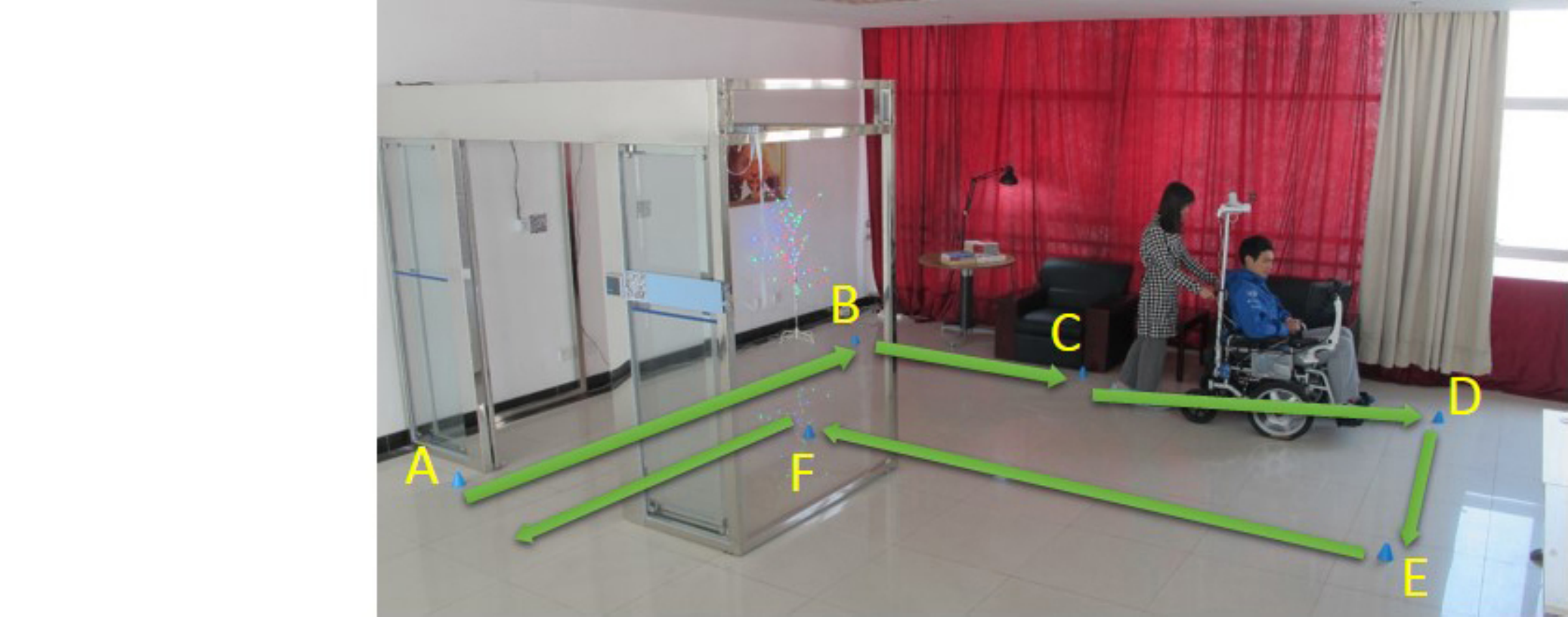}}
\subfigure []
{\includegraphics[width=0.45\textwidth]{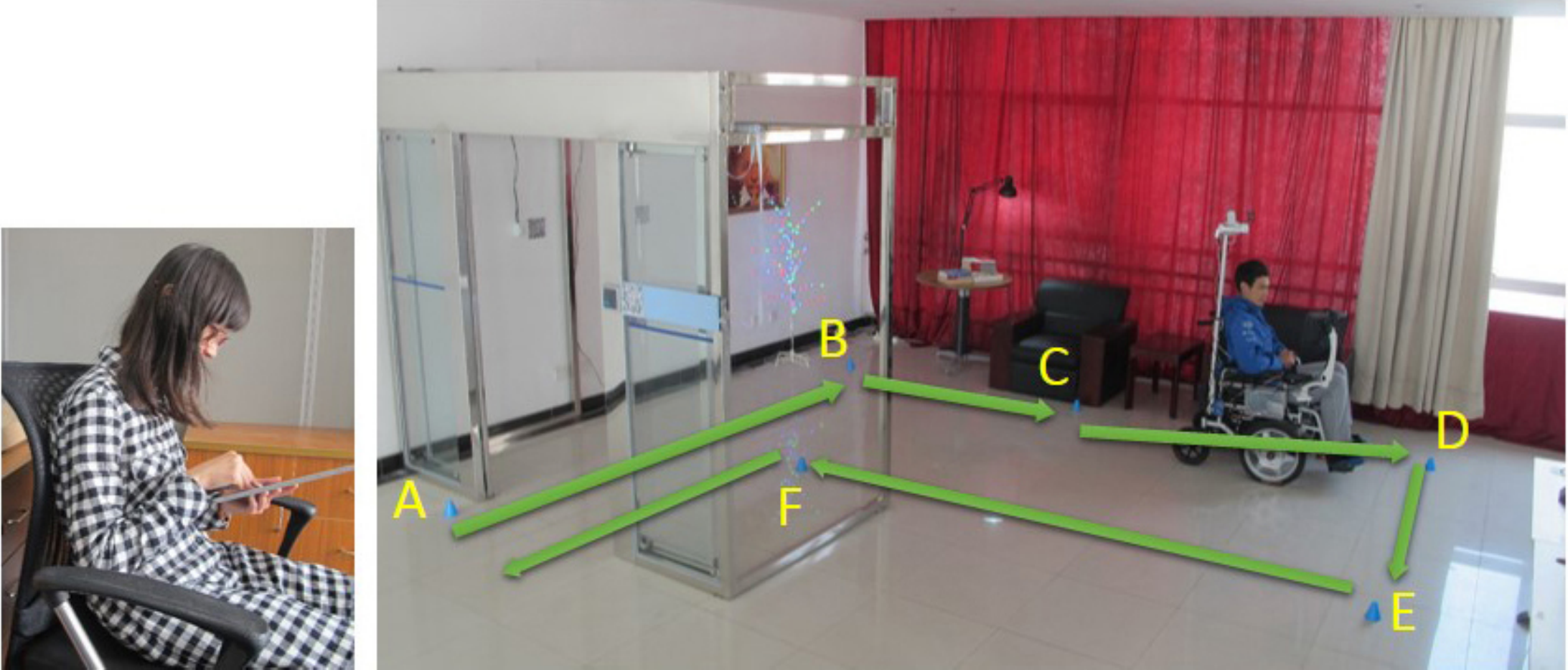}}
\end{center}
\caption{Experiment environment in our lab. (a) One participant sits in the wheelchair and another participant pushes the wheelchair in the presence. (b) One participant sits in the wheelchair and another participant using the pad TIUI to tele-operate the wheelchair in a remote location.}
\label{fig:fig7}
\end{figure}

We constructed the experimental room simulating a living environment, containing sofas, tables, and chairs, as shown in Fig.~\ref{fig:fig7}. The experiment contained two sessions: pushing the wheelchair in person and pushing the wheelchair in a remote space. In the first session, a local participant sat in the wheelchair and a remote participant pushed the wheelchair by its handles in the room (local space) and followed the path: door in(A)-$>$ lamp(B)-$>$ sofa(C)-$>$ window(D)-$>$ desk(E)-$>$ door out(F), the as shown in Fig.~\ref{fig:fig7} (a). After completing the first session, the remote participant moved to another room (remote space) to carry out the second session, and was asked to push the same local participant in the wheelchair using TIUI to follow the same path, as shown in Fig.~\ref{fig:fig7} (b). The time spent for each of the two sessions was recorded.

After completing the two sessions, each participant was asked to fill in a questionnaire about the evaluation for the maneuverability, feedback, feelings of presence, and comfort of our system by comparing two sessions. For example, maneuverability was measured via remote participants agreement with four items on a four-point scale (1 = strongly disagree to 4 = strongly agree), e.g. "I think it is easy to push the wheelchair during this operation". And local participants were asked questions like following: "How do you feel about the comfort of this operation?", they rated their attitudes ranging from 1 (describes very poor) to 4 (describes very well). The score of feelings of presence and feedback described the sensation of the remote participant being in the presence. The face images on the tele-presence wheelchair screen were rated to calculate a score of how participants felt enjoyable during the operation, in terms of the quality of the operator's on-screen expression on a four-point scale (1 =strongly confused to 4 = strongly excited).

\begin{figure}[t]
\setlength{\abovecaptionskip}{-8.px}
\begin{center}
    \includegraphics[width=0.5\textwidth]{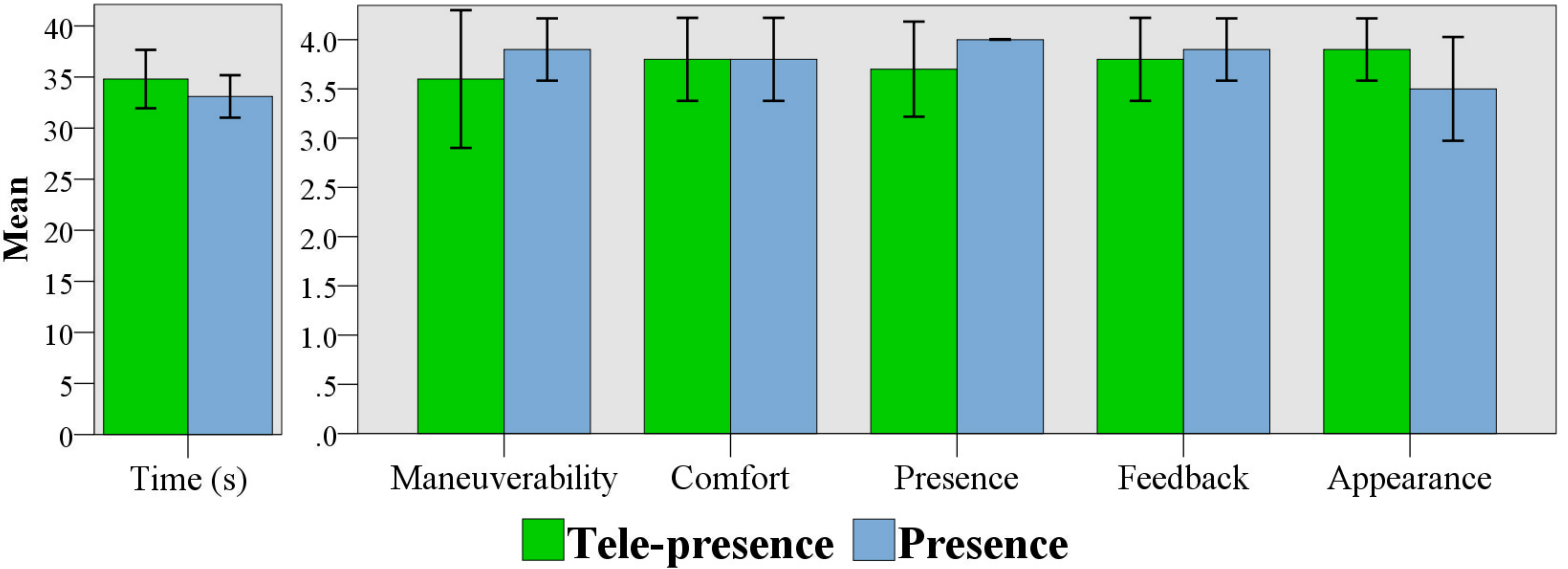}
\end{center}
\caption{Experiment results}
\label{fig:fig8}
\end{figure}

Fig.~\ref{fig:fig8} shows the comparison results. The average time spent for the presence operation (M=33.1, SD=2.079) and the tele-presence operation (M=34.8, SD=2.860) were almost the same. 90\% remote participants deemed that the TIUI was user-friendly and easy to use (M=3.6, SD=0.699). Most of local participants experienced high comfort of the tele-presence wheelchair via the TIUI (M=3.8, SD=0.421) as if remote participants really being in the presence (M=3.7, SD=0.483). What's more, remote participants looked better (M=3.9, SD=0.316) during tele-operation than in the presence (M=3.5, SD=0.527) in terms of the quality of their expression. We can see that participants responded very positively to the tele-presence wheelchair system.

\subsection{Evaluation of navigation system}

We used ten different doors and four kinds of doorway configurations around the office of our lab, which are commonly encountered in daily life, to evaluate the semi-autonomous navigation method. Each configuration is tested with several different offset angles. The four configurations are: A: one doorway in the scene; B: two doorways in the scene; C: three doorways in the scene; D: more than three doorways in the scene.

The evaluation of performance consists of 5 parts: door detection, intention prediction, trajectory generation, trajectory navigation, and the running time (not longer than doing the same task without semi-autonomous navigation). If the 5 parts are all done well, the performance is perfect, if more than 3 parts are done well, the performance is good, and otherwise, the performance is bad. The result is listed in Table 1. It shows that the semi-autonomous navigation method performs well in most cases. With the increase of the doors number, the performance decreases as some doors are not detected.

\begin{figure}[t]
\setlength{\abovecaptionskip}{-8.px}
\begin{center}
    \includegraphics[width=0.45\textwidth]{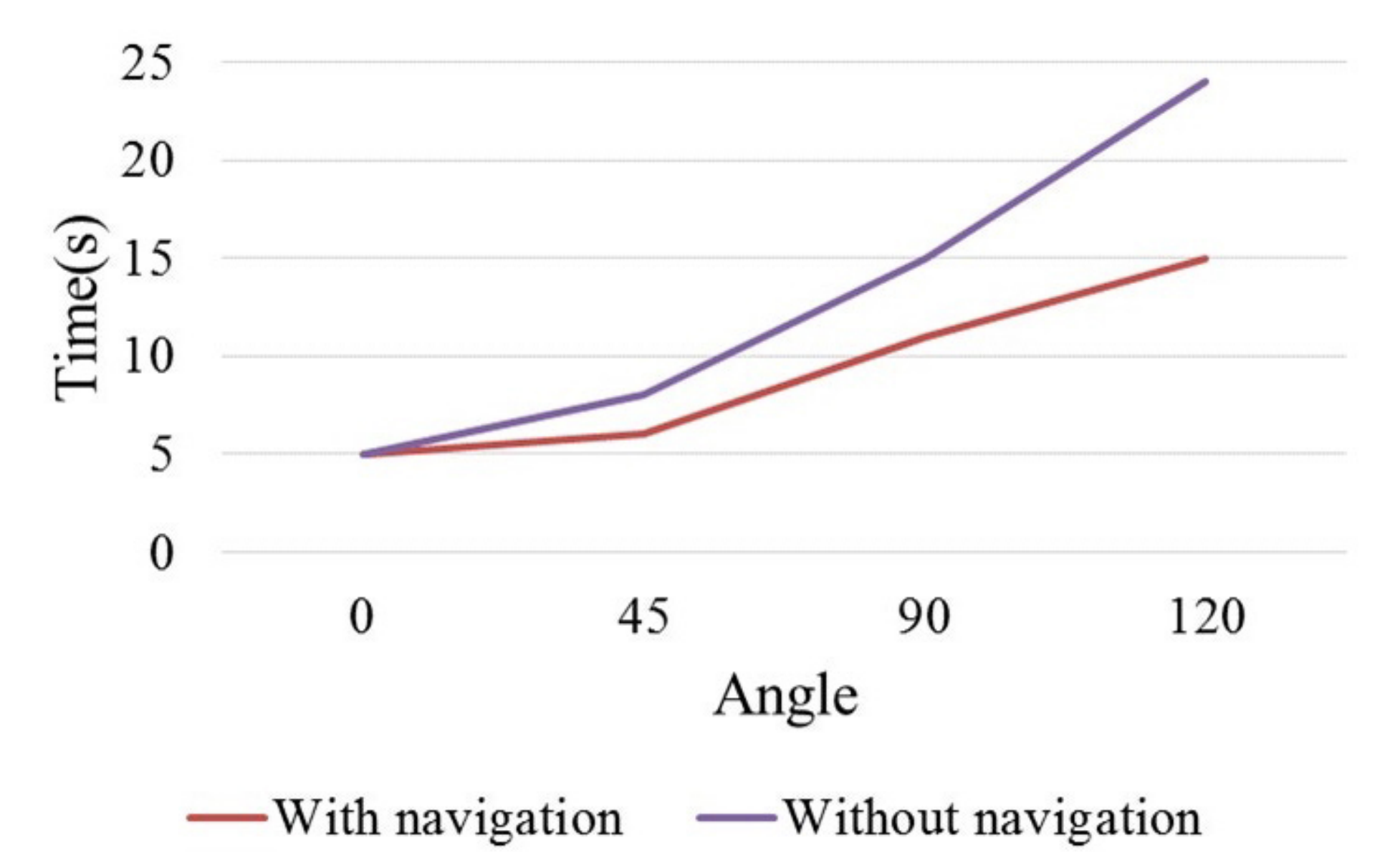}
\end{center}
\caption{The influence of variable offset angles}
\label{fig:fig9}
\end{figure}

Fig.~\ref{fig:fig9} shows the influence of the variable offset angles (angle between the door and the horizontal line) to the performance of system. Time difference is the time saved by using navigation system. When the offset angle is small, the wheelchair can enter the door easily, the difference between with and without semi-autonomous navigation is very small. As the offset angle becomes bigger, the advantage of semi-autonomous navigation is more obvious. When the offset angle is bigger than 90 degree, the time it takes to complete the task is much less than it takes without navigation.

\begin{table}[]
\centering
\caption{THE EVALUATION OF NAVIGATION FOR DIFFERENT SCENES}
\label{my-label}
\begin{tabular}{|c|c|c|c|c|c|c|c|}
\hline
\multirow{2}{*}{Configuration} & \multirow{2}{*}{Rounds} & \multicolumn{2}{c|}{Perfect} & \multicolumn{2}{c|}{Good} & \multicolumn{2}{c|}{Bad} \\ \cline{3-8}
                  &                                   & \#           & \%            & \#         & \%           & \#         & \%          \\ \hline
A                 & 30                                & 26           & 86.6          & 3          & 10.0         & 1          & 3.3         \\ \hline
B                 & 20                                & 15           & 75.0          & 4          & 20.0         & 1          & 5.0         \\ \hline
C                 & 10                                & 6            & 60.0          & 2          & 20.0         & 2          & 20.0        \\ \hline
D                 & 5                                 & 1            & 20.0          & 3          & 60.0         & 1          & 20.0        \\ \hline
\end{tabular}
\end{table}

\section{CONCLUSIONS}

This paper has presented a low-cost wheelchair system which contains an affordable commercial electric wheelchair, an add-on telepresence interaction module, and a new user interface (TIUI). We also described the semi-navigation of the wheelchair to improve tele-operation efficiency and user experience. The experiments show that our system is promising applicable in health care and elder care. The future work includes extending user studies to real environments for promoting usability, and improving semi-navigation algorithm based on live video images to reduce remote operators' overload during teleoperation process.

\addtolength{\textheight}{-12cm}  





\bibliographystyle{IEEEtran}
\bibliography{root}

\end{document}